\documentclass[9pt, conference]{IEEEtran}
\IEEEoverridecommandlockouts
\usepackage{cite}
\usepackage{amsmath,amssymb,amsfonts}
\usepackage{algorithmic}
\usepackage{booktabs}
\usepackage{graphicx}
\usepackage{textcomp}
\usepackage{xcolor}
\usepackage{multirow}
\usepackage{algorithm}
\usepackage{multirow}
\usepackage{amsmath}
\usepackage{hyperref}
\usepackage{balance}
\usepackage{array}
\def\BibTeX{{\rm B\kern-.05em{\sc i\kern-.025em b}\kern-.08em
    T\kern-.1667em\lower.7ex\hbox{E}\kern-.125emX}}

\begin{document}

\title{Speculative Decoding for Verilog: \\Speed and Quality, All in One}


\author{
    \IEEEauthorblockN{
        Changran Xu\textsuperscript{1,3,†}\thanks{† These authors contributed equally.}, 
        Yi Liu\textsuperscript{1,3,†}\thanks{}, 
        Yunhao Zhou\textsuperscript{1,3}, 
        Shan Huang\textsuperscript{2,3}, 
        Ningyi Xu\textsuperscript{2}, 
        and Qiang Xu\textsuperscript{1,3}
    }
    \IEEEauthorblockA{
        \textsuperscript{1}\textit{The Chinese University of Hong Kong},
Shatin, Hong Kong S.A.R.\\
    }
    \IEEEauthorblockA{
        \textsuperscript{2}\textit{Shanghai Jiao Tong University}, Shanghai, China\\
    }
    \IEEEauthorblockA{
        \textsuperscript{3}\textit{National Technology Innovation Center for EDA}, Nanjing, Jiangsu, China\\
    }
    
}

\maketitle

\begin{abstract}

The rapid advancement of large language models (LLMs) has revolutionized code generation tasks across various programming languages. However, the unique characteristics of programming languages, particularly those like Verilog with specific syntax and lower representation in training datasets, pose significant challenges for conventional tokenization and decoding approaches. In this paper, we introduce a novel application of speculative decoding for Verilog code generation, showing that it can improve both inference speed and output quality, effectively achieving speed and quality all in one.

Unlike standard LLM tokenization schemes, which often fragment meaningful code structures, our approach aligns decoding stops with syntactically significant tokens,
making it easier for models to learn the token distribution. This refinement addresses inherent tokenization issues and enhances the model’s ability to capture Verilog’s logical constructs more effectively.
Our experimental results show that our method achieves up to a 5.05× speedup in Verilog code generation and increases pass@10 functional accuracy on RTLLM by up to 17.19\% compared to conventional training strategies.
These findings highlight speculative decoding as a promising approach to bridge the quality gap in code generation for specialized programming languages.

\end{abstract}


\section{Introduction} \label{sec:introduction}


The rapid advancement in large language models (LLMs) has transformed code generation across various programming languages~\cite{wang2021codet5,wang2023codet5+,roziere2023code,zhu2024deepseek}. These models, driven by advanced pre-training techniques, have achieved notable success in generating syntactically and semantically correct code for widely used languages like Python and C++. However, their application to specialized languages such as Verilog, which is critical for hardware design and verification, remains limited due to the unique challenges posed by Verilog’s syntax intricacies and its underrepresentation in training datasets.


Unlike natural languages, programming languages like Verilog are governed by strict syntactic rules, where even minor deviations can lead to significant compilation failures or functional errors. Traditional tokenization methods, such as Byte Pair Encoding (BPE)~\cite{sennrich2015neural}, fragment meaningful code structures into subword units, obscuring the logical relationships inherent in Verilog. Consequently, models often struggle to accurately capture the syntax and semantics of Verilog code. Grammar-based approaches attempt to mitigate this issue by representing code as sequences of grammar rules~\cite{rabinovich2017abstract,sun2019grammar,sun2020treegen,xiong2022l2s,zhu2022grape,zhu2024grammart5}. While effective for smaller models and datasets, these methods face scalability issues when applied to LLMs, due to vocabulary explosion and distribution shifts. Moreover, these methods remain largely unexplored for underrepresented languages like Verilog, which presents an additional layer of complexity.

\begin{figure}[ht]
    \centering
    \includegraphics[width=\linewidth]{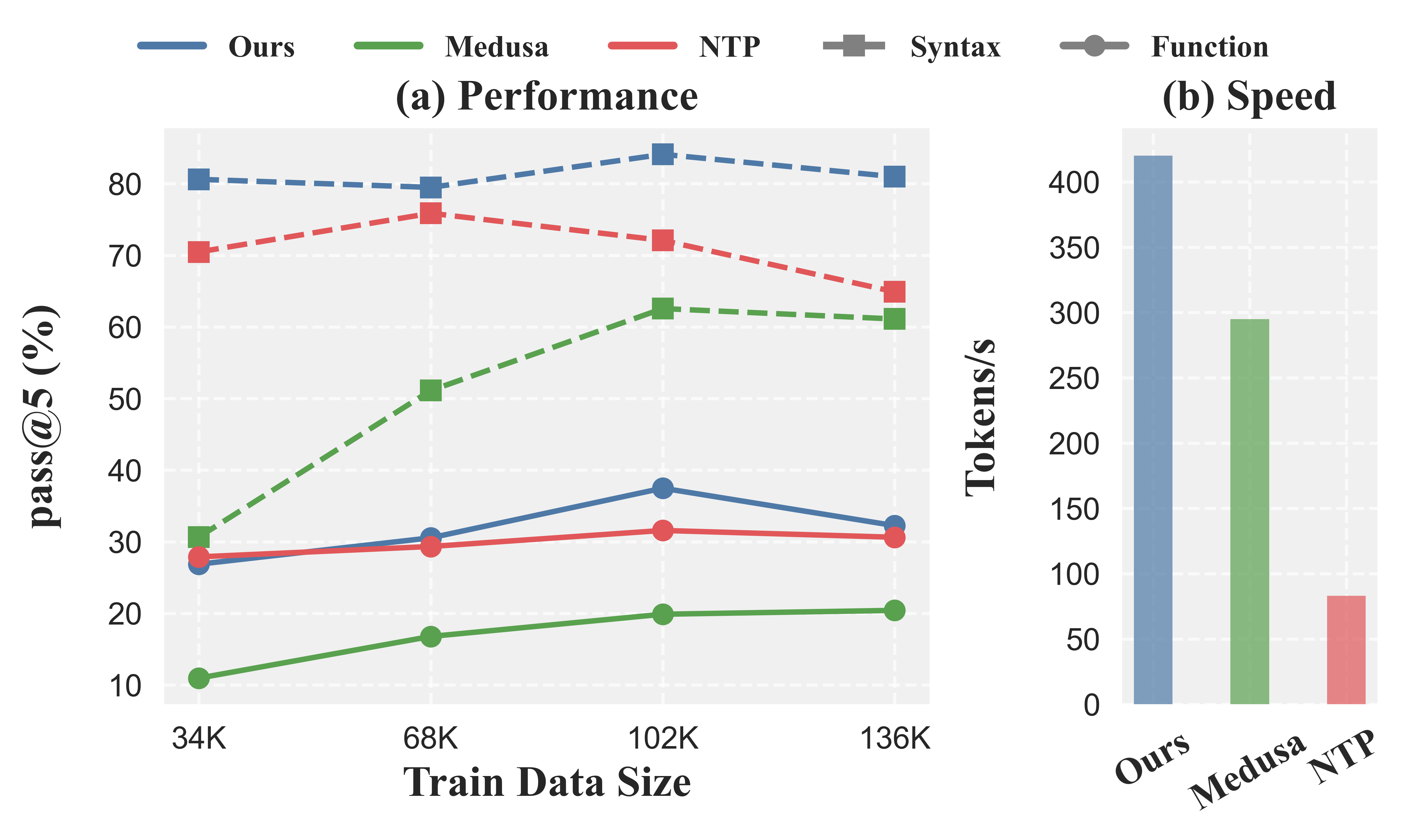}
    \vspace{-13pt}
    \caption{A brief comparison of the performance and speed of our method against the M\textsc{edusa} method and the conventional next token prediction (NTP) approach. The experiments are conducted using the CodeLlama-7b model, with performance metrics evaluated on the RTLLM benchmark.}
    \vspace{-6pt}
    \label{fig:brief_compare}
\end{figure}

To address these challenges, we propose a novel application of speculative decoding~\cite{leviathan2023fast,chen2023accelerating,miao2024specinfer,cai2024medusa} for Verilog code generation. 
While speculative decoding has traditionally been employed to accelerate LLM inference by predicting multiple tokens simultaneously to reduce latency, our key insight is that \emph{speculative decoding, when carefully aligned with syntactically significant tokens, can also improve the quality of code generation}.
By structuring decoding stops around meaningful fragments in Verilog, such as identifiers, keywords, and operators,  we enable the model to better capture the logical and structural relationships inherent in the language. This alignment simplifies the learning of Verilog's token distribution, mitigating issues caused by conventional tokenization schemes and enhancing the syntactic and functional correctness of the generated code. Specifically, the main contributions of our work include:

\begin{itemize}
    \item To the best of our knowledge, this is the first application of speculative decoding aimed not only at improving inference speed but also enhancing output quality.  
    \item We propose a simple yet effective method for identifying syntactically significant tokens in Verilog using abstract syntax trees (ASTs) and modify the speculative decoding scheme by incorporating syntax-enriched labels to align decoding stops with these tokens.
    \item Our decoding scheme further unlocks the potential of the original M\textsc{edusa} method~\cite{cai2024medusa} by training it with dynamic labels in a flexible manner, increasing the number of effective heads and further accelerating the speedup.
\end{itemize}

We conduct experiments with two models, CodeT5p-220m-bimodal~\cite{wang2023codet5+} and CodeLlama-7b~\cite{roziere2023code}. The results show that our method achieves up to a 5.05× speedup in Verilog code generation and increases pass@10 functional accuracy on RTLLM~\cite{lu2024rtllm} by up to 17.19\% compared to conventional training strategies. Impressively, our approach also accelerates model inference by 1.42–2.29× over the original M\textsc{edusa} method, highlighting its ability to deliver both speed and quality simultaneously.

\section{Related Work} \label{sec:related_work}

\subsection{LLM for Verilog Generation}

Recent advancements in LLMs have shown significant potential for automating the generation of Verilog code from high-level prompts~\cite{pearce2020dave,blocklove2023chip,thakur2023benchmarking,liu2023verilogeval,chang2023chipgpt}, demonstrating their ability to address the design challenges faced by hardware developers.
While some works~\cite{thakur2024verigen,liu2024rtlcoder,chang2024data,zhang2024mg,cui2024origen,zhao2024codev} have fine-tuned open-source models to improve Verilog code generation, they often treat Verilog code as if it were natural language, using tokenizers originally designed for natural language models without modification. This approach often leads to the fragmentation of meaningful Verilog code structures, which hinders the model's ability to accurately capture the syntactic structure of Verilog and generate syntactically correct code.
Furthermore, the scarcity of Verilog code datasets exacerbates the challenge of effectively learning Verilog's syntax.
To address this issue, we propose a novel application of speculative decoding for Verilog code generation. 
By integrating conventional tokenization with Verilog’s syntactical structure, our method allows the model to generate more accurate Verilog code more efficiently.

\subsection{Syntax-Aware Tokenization for Code}

Driven by recent developments in LLMs, a variety of pretrained models for programming languages have emerged~\cite{wang2021codet5,wang2023codet5+,roziere2023code,zhu2024deepseek,thakur2024verigen,liu2024rtlcoder,chang2024data,zhang2024mg,cui2024origen,zhao2024codev}. However, most of these models represent code as token sequences using the BPE algorithm for pre-training, which often hinders their ability to capture code's syntactic structure and fails to ensure the syntactic correctness of the generated output.
To address this limitation, some models~\cite{wang2021syncobert,jiang2021treebert,guo2022unixcoder} have attempted to explicitly incorporate syntax by representing code as AST sequences, which are obtained by traversing the AST in pre-order and recording the symbol of each node.
However, to maintain the tree structure in the node sequences, these approaches introduce additional nodes, which significantly increase sequence length and GPU memory usage, potentially compromising model performance. Furthermore, these models only leverage the AST representation in the encoder and do not ensure syntactic correctness during code generation.
Meanwhile, several non-pretrained code generation models~\cite{rabinovich2017abstract,sun2019grammar,sun2020treegen,xiong2022l2s,zhu2022grape} have used grammar rule sequences to represent code. Specifically, they traverse the AST in pre-order and record the grammar rules used to expand each non-terminal.
The integration of grammar rules has been shown to improve performance. However, these models have only been tested in non-pretrained settings and do not involve LLMs.
When applied to pre-training with large code corpora, these models face a significant challenge due to the \textit{big vocabulary} problem, which arises from the large number of user-defined identifiers and could degrade the model's performance.
In contrast, BPE is designed to find a relatively small set of subtokens whose concatenations could represent a large token set.
To integrate BPE with grammars, GrammarT5~\cite{zhu2024grammart5} uses a variant of grammar rule sequence to represent code, called Tokenized Grammar Rule Sequence (TGRS).
Despite this innovation, GrammarT5 still faces generalization issues when applied to token sequences and has only demonstrated marginal improvements in smaller models, leaving its effectiveness on larger models unexplored.
Moreover, all these methods have ignored programming languages like Verilog, which are underrepresented in training datasets, making these challenges even more significant. 
To overcome these limitations, we propose a more natural way to integrates BPE with grammar without significantly altering the distribution of code tokens.
By introducing a novel application of speculative decoding for Verilog, our method can enhance both the speed and quality of Verilog code generation.

\subsection{Speculative Decoding}

As the size of language models continues to grow, inference latency has become a critical challenge for practical applications. 
To alleviate this issue, speculative decoding~\cite{leviathan2023fast,chen2023accelerating,miao2024specinfer} has been proposed to reduce the number of decoding steps. 
Traditional speculative decoding employs a smaller draft model to generate an initial token sequence, which the original, larger model then refines to produce an acceptable continuation.
However, it remains challenging to acquire and maintain a separate draft model.
To address this, M\textsc{edusa}~\cite{cai2024medusa} introduces additional decoding heads that predict multiple tokens concurrently, enabling seamlessly integration into existing LLM systems. 
M\textsc{edusa} offers two fine-tuning approaches tailored to different use cases: 
(1) M\textsc{edusa}-1, which is fine-tuned on a frozen LLM, ensuring lossless acceleration; and (2) M\textsc{edusa}-2, which is jointly fine-tuned with the backbone LLM, achieving higher prediction accuracy and greater speedups.
While originally developed for accelerating LLM inference, we find that speculative decoding can also preserve the syntactic structure of code by modifying its decoding mechanisms. By aligning decoding stops with syntactically significant tokens, speculative decoding enhances both inference speed and output quality in Verilog code generation. Fig.~\ref{fig:brief_compare} presents a brief comparison of the performance and speed of our method against the original M\textsc{edusa} method and the conventional next token prediction (NTP) approach.
\section{Methodology} \label{sec:method}

\begin{figure}[ht]
    \centering
    \includegraphics[width=1.0\linewidth]{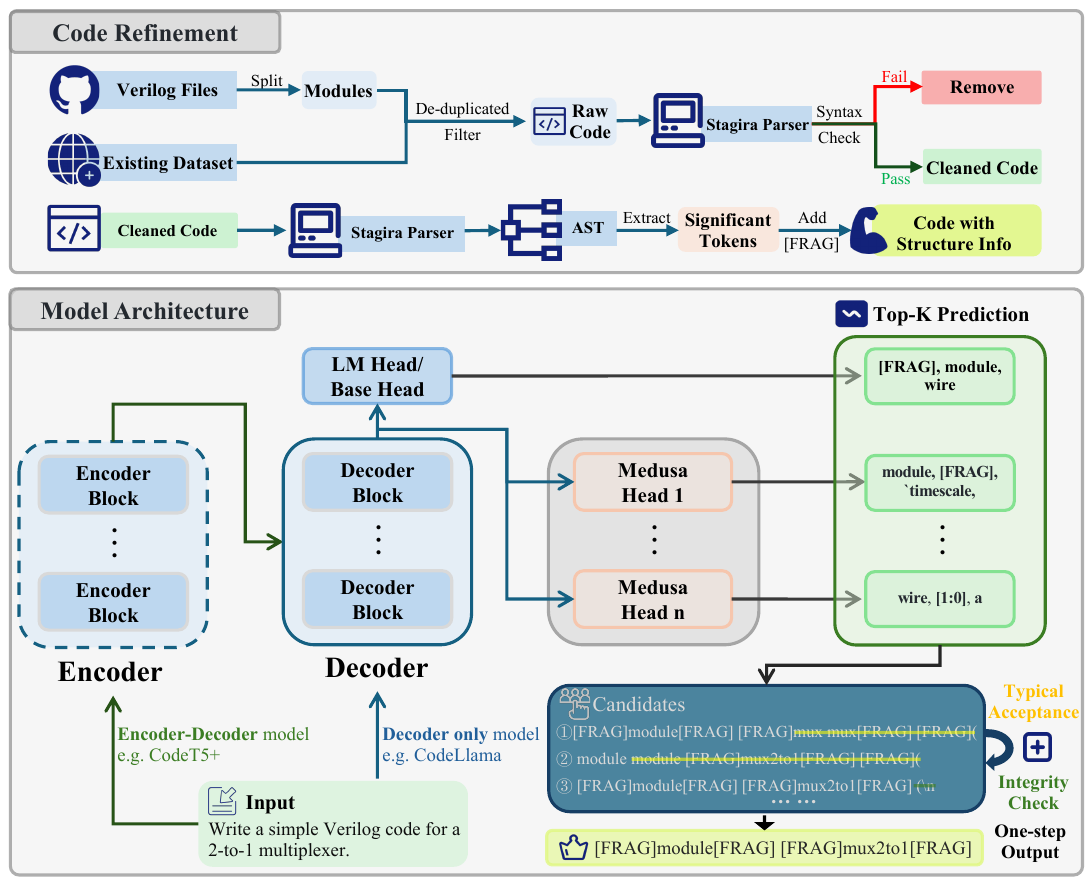}
    \caption{The overview of the data refinement process and model architecture.}
    \vspace{-6pt}
    \label{fig:code_refinement_and_model_architecture}
\end{figure}

\subsection{Dataset Construction}
\label{subsection:dataset_construction}

Our dataset comprises \texttt{.v} files collected from Github using \texttt{Verilog} as the search keyword. Each file is segmented into functional Verilog modules, and duplicates are removed using MinHash and Jaccard similarity metrics~\cite{yan2017privmin}. We also filter out files lacking complete \texttt{module} and \texttt{endmodule} structures or primarily consisting of comments. Additionally, we supplement our dataset with data from the open-source projects MG-Verilog~\cite{zhang2024mg} and RTLCoder~\cite{liu2024rtlcoder}, resulting a total of 13,6134 data items.
To ensure the quality of the dataset, we apply the Stagira Verilog parser~\cite{chen2023incremental} to perform syntax checks on all code samples and retain only those that pass (\textit{i.e.}, cleaned code). For these cleaned code samples, we use the parser to generate their corresponding ASTs, from which the syntactically significant tokens are extracted. The entire code refinement process is illustrated in Fig.~\ref{fig:code_refinement_and_model_architecture}.
For the cleaned code collected from Github, we leverage GPT-4~\cite{achiam2023gpt} to generate functional descriptions, while the data from MG-Verilog and RTLCoder already include code summaries.

\subsection{Model Architecture}
\label{subsection:model_architecture}

In this work, we evaluate our approach using two base models: CodeT5p-220m-bimodal~\cite{wang2023codet5+} and CodeLlama-7b~\cite{roziere2023code}, which differ in architecture and size.
Specifically, CodeT5p employs an encoder-decoder architecture while CodeLlama is a decoder-only model. 
As shown in Fig.~\ref{fig:code_refinement_and_model_architecture}, we augment the base models with additional heads attached to their last hidden states to predict multiple tokens concurrently, following the original M\textsc{edusa} method~\cite{cai2024medusa}.
To improve both inference speed and the output quality of Verilog code generation, we modify M\textsc{edusa}'s decoding mechanism to align decoding stops with syntactically significant tokens, which ensures that the model maintains a complete syntactic structure at each decoding step, enabling it to explicitly capture the Verilog syntax.
For detailed explanations, please refer to Section~\ref{subsection:special_token}.


During the decoding process, given that the model is currently at position $t$, the base model will predict the token at $t+1$, while the $i$-th head will predict the token at $t+i+1$. Each token generated by the heads is first evaluated by the typical acceptance strategy~\cite{cai2024medusa}:
\begin{align}
p_{\text{base}}(x_{t+i+1}|x_1, x_2, \dots, x_{t+i}) > \nonumber \\
\min\left(\epsilon, \delta \exp\left(-H\left(p_{\text{base}}(\cdot|x_1, x_2, \dots, x_{t+i})\right)\right)\right) \label{eq:typical_acceptance}
\end{align}
where $p_{\text{base}}$ represents the prediction probability of the base model, $H(\cdot)$ denotes the entropy function, and $\epsilon$, $\delta$ are hyper-parameters. 
A token is accepted only if~\eqref{eq:typical_acceptance} holds for it and all preceding tokens.
At this decoding step, suppose only the tokens predicted by the first $u$ heads are accepted. 
These tokens, spanning positions $t+1$ to $t+u+1$, are then re-evaluated to ensure they form a complete syntactic structure.
Any extraneous tokens that compromise the integrity of the code fragment are discarded. For instance, if the tokens from $t+1$ to $t+v+1$ (where $v < u$) already constitute a complete code fragment, such as an identifier or keyword, the outputs from the remaining heads, corresponding to $t+v+2$ to $t+u+1$, are discarded.
During decoding, we maintain several candidates comprising the top-$k$ predictions from the base model and additional heads. The final prediction for the current step is the \emph{longest accepted prefix} among all candidates.

\subsection{Syntax-Enriched Labels}
\label{subsection:special_token}

\begin{figure}[ht]
    \centering
    \includegraphics[width=1.0\linewidth]{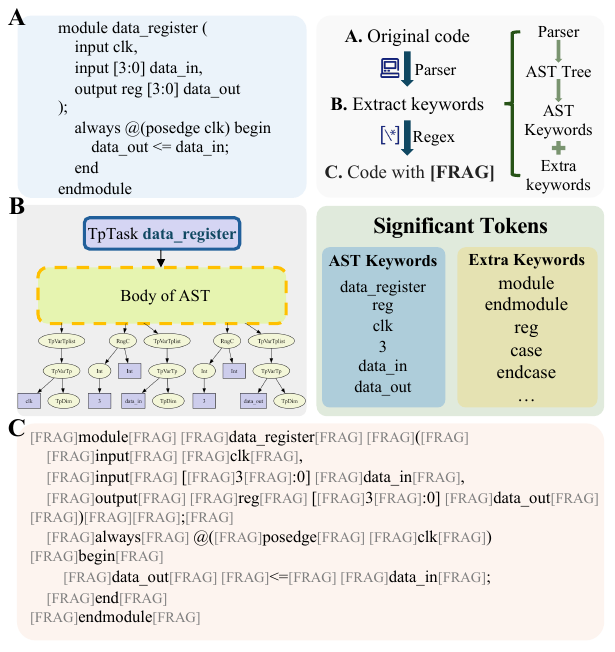}
    \caption{An example demonstrating the identification and extraction of syntactically significant tokens from Verilog code.}
    \vspace{-6pt}
    \label{fig:syntactically_significant_tokens}
\end{figure}

In this section, we explain how syntactically significant tokens are identified and present a novel method to construct syntax-enriched labels for incorporating syntax into the speculative decoding process.
To identify syntactically significant tokens, we first parse Verilog code into ASTs, from which we extract leaf nodes and non-terminal nodes that contains critical information as keywords. Additionally, we supplement these keywords with commonly used Verilog constructs, such as \texttt{negedge} and \texttt{endmodule}. Together, these keywords form the syntactically significant tokens.
We then use regular expressions (regex) to segment the code into meaningful fragments that maintain syntax integrity based on these keywords. At each segmentation point, we insert a special token \texttt{[FRAG]} to prepare for constructing syntax-enriched labels. 
Fig.~\ref{fig:syntactically_significant_tokens} illustrates this process with an example.

\begin{figure*}[ht]
    \centering
    \includegraphics[width=0.712\textwidth]{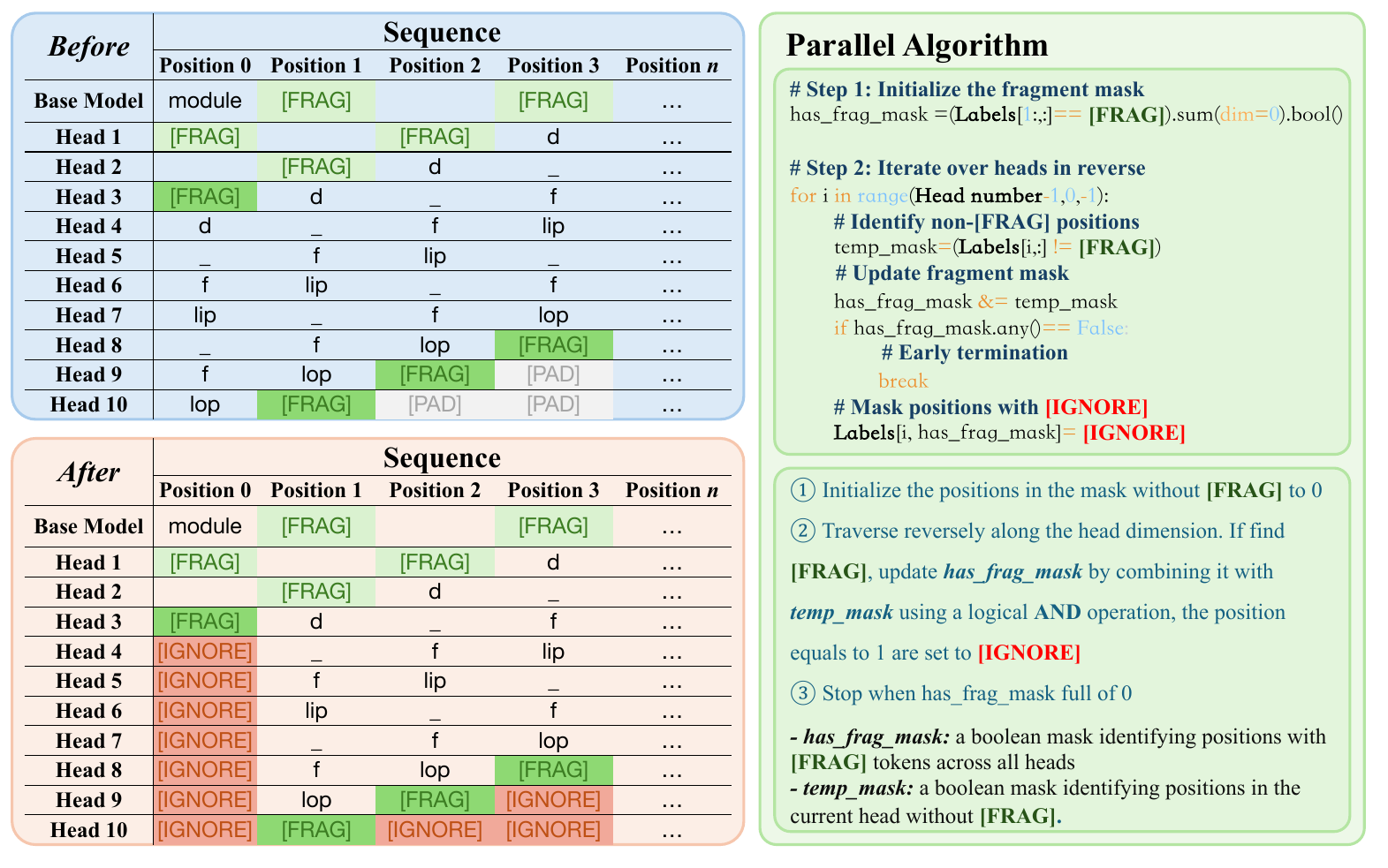}
    \caption{The construction of syntax-enriched labels for aligning decoding stops with syntactically significant tokens. The top-left panel illustrates the initial labels of Verilog code filled with \texttt{[FRAG]} tokens, while the bottom-left panel depicts the final syntax-enriched labels used for training. The right panel presents the parallel algorithm for accelerating the label construction process.}
    \label{fig:label_pre}
\end{figure*}

To incorporate syntax information into the speculative decoding process, we modify the decoding mechanism by introducing syntax-enriched labels.
Specifically, the label for the base model, denoted as $L_{0}$ with a length of $sequence\_length$, is the tokenized version of the Verilog code filled with \texttt{[FRAG]} tokens.
And the label for the $i$-th head is first derived by left-shifting the base model’s label by $i$, resulting in $L_{i}=L_{0}[i:]$. \texttt{[PAD]} tokens are then appended to ensure that the head labels align in length with the base model’s label. The labels for all heads are then concatenated with the base model’s label, forming the combined label $Labels$ with size $(num\_heads+1) \times sequence\_length$. 
As shown in Fig.~\ref{fig:label_pre}, for each sequence position, we identify the position of the last \texttt{[FRAG]} token along the decoding heads. Labels beyond this position are replaced with \texttt{[IGNORE]} tokens, excluding them from loss computation. This ensures that, for any given sequence position $s$, the base model’s label and the corresponding head labels, $Labels\left[:, s\right]$, represent syntactically meaningful fragments by excluding incomplete information. 
The syntax-enriched labels align decoding stops with syntactically significant tokens, enabling the model to better capture the syntactic structure of Verilog code.
Additionally, the progressive increase of the proportion of \texttt{[IGNORE]} tokens in the labels of later heads reduces their prediction difficulty, improving the model’s training efficiency and allowing us to train more robust heads than the original M\textsc{edusa} method, which further accelerates inference. To optimize performance, we design a parallel algorithm that fully parallelizes the procedure, significantly speeding up the process and minimizing computational overhead during training. The pseudo-code for this algorithm is provided in Fig.~\ref{fig:label_pre}.



\section{Experiments} \label{sec:experiment}


\subsection{Experimental Setup}

We select CodeLlama-7b-Instruct-hf~\cite{roziere2023code} (CodeLlama) and CodeT5p-220m-bimodal~\cite{wang2023codet5+} (CodeT5p) as our base models. For each base model, we append 10 additional heads and fine-tune it for Verilog code generation, comparing our syntax-enriched training method (Ours) with the original M\textsc{edusa}-2 method~\cite{cai2024medusa} (Medusa) as one baseline. We also consider the conventional NTP scheme (NTP) as another baseline.

\subsubsection{Training Data}

The training data is constructed using the dataset introduced in Section~\ref{subsection:dataset_construction}. Specifically, it is formatted into the Alpaca style~\cite{alpaca}, with the natural language description as input and the corresponding Verilog code as output, resulting in 136K samples for fine-tuning.
Due to the 2048-token limit of the CodeT5p model, we exclude examples exceeding this threshold, resulting in 128K training samples for models based on the CodeT5p architecture. The data used for the two baseline methods is identical to ours, except that it does not incorporate the syntax-enriched labels.
To evaluate model performance across varying training data sizes, we fine-tune the models not only on the full dataset but also on random subsets comprising 1/4, 1/2, and 3/4 of the original data.

\subsubsection{Model Training}

All model training is conducted on four NVIDIA A800-SXM4-80GB GPUs using the Distributed Data Parallel (DDP) module from PyTorch.

We utilize the Axolotl framework~\cite{axolotl2023} to fine-tune models based on CodeLlama with QLoRA~\cite{dettmers2023qlora}, using consistent hyperparameters: a LoRA adapter rank of 32, $\alpha$ set to 16, and a dropout rate of 0.05. Additional hyperparameters are also kept constant, including a micro-batch size of 1, an 8-bit AdamW optimizer with a cosine learning rate scheduler, an initial learning rate of $5e^{-4}$ for the base model, a warmup period of 40 steps, and a maximum sequence length of 8192 tokens. To improve training throughput, we employ the multipack method, which combines multiple short sequences into a single batch.
For models based on CodeT5p, we fine-tune directly using a batch size of 2, the AdamW optimizer, a learning rate of $5e^{-4}$ for the base model, a warmup ratio of 0.1, and a maximum sequence length of 2048 tokens. 

For models with additional heads, the learning rate for the decoding heads is set to four times that of the base model. 
The overall loss for these multi-head models is computed using the method proposed in M\textsc{edusa}~\cite{cai2024medusa}:

\begin{equation}
\text{Loss} = Loss_{base} + \lambda\cdot\sum_{i=1}^{n} (Loss_{\text{head}_i} \cdot \gamma^i)
\label{eq:loss}
\end{equation}
In this equation, \(\lambda\) is a dynamic weighting factor that adjusts the influence of the heads' losses on the overall loss. During training, \(\lambda\) follows a sine growth pattern, increasing from 0 to 0.2 as training progresses. The parameter \(\gamma\), set to 0.8 in our experiments, serves as a decay coefficient to differentially weight each head’s loss. Here, $n$ represents the number of heads, fixed at 10 in our experiments.

\subsubsection{Model Inference}

\begin{figure*}[ht]
    \centering
    \includegraphics[width=0.68\linewidth]{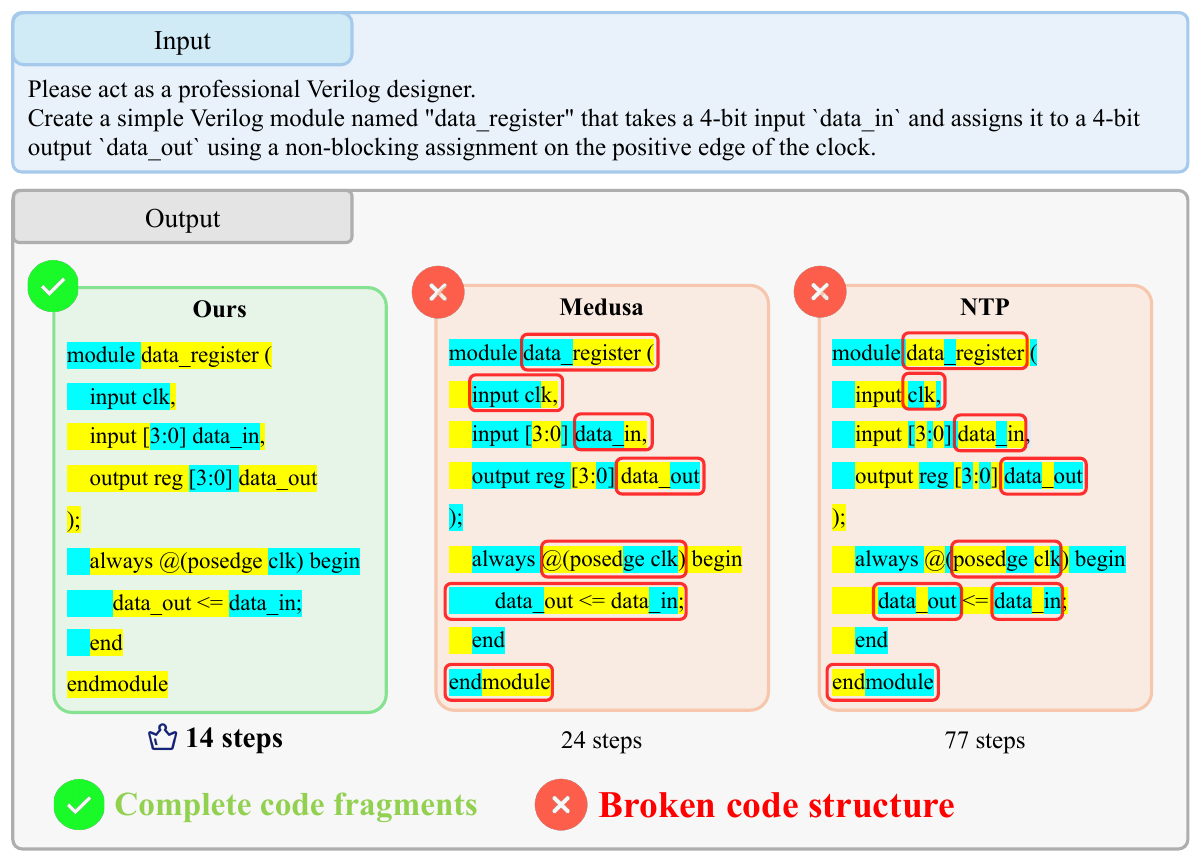}
    \caption{The comparison of the decoding processes for a specific example using our method, Medusa, and NTP. Remarkably, our method generates the output in significantly fewer steps while preserving the integrity of syntactic structure at each decoding step.}
    \label{fig:output_example}
\end{figure*}

During inference, all models are loaded in float16 format. CodeLlama-based models are configured with a maximum token length of 8192, while CodeT5p-based models are limited to 2048 tokens. For speed evaluation, each prompt is processed using two decoding methods: greedy decoding and sampling decoding at a temperature of 0.8. For quality evaluation of the generated Verilog code, 20 responses are sampled per prompt at temperatures of 0.2, 0.4, 0.6, and 0.8. The final result for each prompt is determined by selecting the output with the highest accuracy across all temperatures.

\begin{table*}
\centering
\caption{Evaluation results for the quality of generated Verilog code}
\resizebox{\textwidth}{!}{%
\begin{tabular}{|c|c|c|c|ccc|ccc|ccc|ccc|}
\hline
\multirow{2}{*}{Test} & \multirow{2}{*}{Model} & \multirow{2}{*}{Data Size} & \multirow{2}{*}{Benchmark} & \multicolumn{3}{c|}{pass@1 (\%)} & \multicolumn{3}{c|}{pass@5 (\%)} & \multicolumn{3}{c|}{pass@10 (\%)} & \multicolumn{3}{c|}{Pass Rate (\%)} \\ \cline{5-16} 
 &  &  &  & \multicolumn{1}{c|}{Ours} & \multicolumn{1}{c|}{Medusa} & NTP & \multicolumn{1}{c|}{Ours} & \multicolumn{1}{c|}{Medusa} & NTP & \multicolumn{1}{c|}{Ours} & \multicolumn{1}{c|}{Medusa} & NTP & \multicolumn{1}{c|}{Ours} & \multicolumn{1}{c|}{Medusa} & NTP \\ \hline
\multirow{16}{*}{Function} & \multirow{8}{*}{CodeLlama} & \multirow{2}{*}{34K} & RTLLM & \multicolumn{1}{c|}{16.21} & \multicolumn{1}{c|}{4.66} & 16.72 & \multicolumn{1}{c|}{26.87} & \multicolumn{1}{c|}{10.96} & 27.89 & \multicolumn{1}{c|}{33.85} & \multicolumn{1}{c|}{13.07} & 32.62 & \multicolumn{1}{c|}{41.38} & \multicolumn{1}{c|}{13.79} & 37.93 \\ \cline{4-16} 
 &  &  & VGen & \multicolumn{1}{c|}{30.59} & \multicolumn{1}{c|}{27.06} & 29.12 & \multicolumn{1}{c|}{49.17} & \multicolumn{1}{c|}{34.48} & 47.65 & \multicolumn{1}{c|}{56.78} & \multicolumn{1}{c|}{38.04} & 55.16 & \multicolumn{1}{c|}{64.71} & \multicolumn{1}{c|}{41.18} & 58.82 \\ \cline{3-16} 
 &  & \multirow{2}{*}{68K} & RTLLM & \multicolumn{1}{c|}{18.28} & \multicolumn{1}{c|}{13.28} & 18.79 & \multicolumn{1}{c|}{30.55} & \multicolumn{1}{c|}{16.78} & 29.34 & \multicolumn{1}{c|}{35.52} & \multicolumn{1}{c|}{19.87} & 34.24 & \multicolumn{1}{c|}{41.38} & \multicolumn{1}{c|}{24.14} & 37.93 \\ \cline{4-16} 
 &  &  & VGen & \multicolumn{1}{c|}{32.06} & \multicolumn{1}{c|}{25.59} & 24.12 & \multicolumn{1}{c|}{47.01} & \multicolumn{1}{c|}{32.88} & 44.12 & \multicolumn{1}{c|}{53.61} & \multicolumn{1}{c|}{36.59} & 51.81 & \multicolumn{1}{c|}{58.82} & \multicolumn{1}{c|}{41.18} & 58.82 \\ \cline{3-16} 
 &  & \multirow{2}{*}{102K} & RTLLM & \multicolumn{1}{c|}{20.52} & \multicolumn{1}{c|}{13.10} & 17.07 & \multicolumn{1}{c|}{\textbf{37.48}} & \multicolumn{1}{c|}{19.88} & 31.58 & \multicolumn{1}{c|}{\textbf{46.29}} & \multicolumn{1}{c|}{23.50} & 38.25 & \multicolumn{1}{c|}{\textbf{55.17}} & \multicolumn{1}{c|}{27.59} & 41.38 \\ \cline{4-16} 
 &  &  & VGen & \multicolumn{1}{c|}{31.18} & \multicolumn{1}{c|}{26.76} & 32.35 & \multicolumn{1}{c|}{53.42} & \multicolumn{1}{c|}{34.36} & 53.82 & \multicolumn{1}{c|}{63.17} & \multicolumn{1}{c|}{36.81} & 61.72 & \multicolumn{1}{c|}{70.59} & \multicolumn{1}{c|}{41.18} & 64.71 \\ \cline{3-16} 
 &  & \multirow{2}{*}{136K} & RTLLM & \multicolumn{1}{c|}{\textbf{21.55}} & \multicolumn{1}{c|}{13.79} & 12.24 & \multicolumn{1}{c|}{32.25} & \multicolumn{1}{c|}{20.42} & 30.63 & \multicolumn{1}{c|}{38.56} & \multicolumn{1}{c|}{25.04} & 37.55 & \multicolumn{1}{c|}{44.83} & \multicolumn{1}{c|}{34.48} & 37.93 \\ \cline{4-16} 
 &  &  & VGen & \multicolumn{1}{c|}{\textbf{34.12}} & \multicolumn{1}{c|}{22.35} & 31.76 & \multicolumn{1}{c|}{\textbf{55.47}} & \multicolumn{1}{c|}{32.78} & 51.64 & \multicolumn{1}{c|}{\textbf{65.51}} & \multicolumn{1}{c|}{35.01} & 63.21 & \multicolumn{1}{c|}{\textbf{76.47}} & \multicolumn{1}{c|}{35.29} & \textbf{76.47} \\ \cline{2-16} 
 & \multirow{8}{*}{CodeT5p} & \multirow{2}{*}{32K} & RTLLM & \multicolumn{1}{c|}{1.21} & \multicolumn{1}{c|}{0.34} & 0.00 & \multicolumn{1}{c|}{3.87} & \multicolumn{1}{c|}{1.54} & 0.00 & \multicolumn{1}{c|}{5.15} & \multicolumn{1}{c|}{2.63} & 0.00 & \multicolumn{1}{c|}{6.90} & \multicolumn{1}{c|}{3.45} & 0.00 \\ \cline{4-16} 
 &  &  & VGen & \multicolumn{1}{c|}{14.41} & \multicolumn{1}{c|}{0.29} & 1.76 & \multicolumn{1}{c|}{19.79} & \multicolumn{1}{c|}{1.47} & 6.48 & \multicolumn{1}{c|}{19.97} & \multicolumn{1}{c|}{2.94} & 11.15 & \multicolumn{1}{c|}{23.53} & \multicolumn{1}{c|}{5.88} & 17.65 \\ \cline{3-16} 
 &  & \multirow{2}{*}{64K} & RTLLM & \multicolumn{1}{c|}{0.86} & \multicolumn{1}{c|}{1.90} & 0.52 & \multicolumn{1}{c|}{3.95} & \multicolumn{1}{c|}{4.89} & 2.07 & \multicolumn{1}{c|}{6.99} & \multicolumn{1}{c|}{6.08} & 3.09 & \multicolumn{1}{c|}{10.34} & \multicolumn{1}{c|}{6.90} & 3.45 \\ \cline{4-16} 
 &  &  & VGen & \multicolumn{1}{c|}{\textbf{15.88}} & \multicolumn{1}{c|}{10.59} & 7.06 & \multicolumn{1}{c|}{21.09} & \multicolumn{1}{c|}{13.23} & 13.54 & \multicolumn{1}{c|}{22.91} & \multicolumn{1}{c|}{14.71} & 16.22 & \multicolumn{1}{c|}{23.53} & \multicolumn{1}{c|}{17.65} & 17.65 \\ \cline{3-16} 
 &  & \multirow{2}{*}{96K} & RTLLM & \multicolumn{1}{c|}{5.00} & \multicolumn{1}{c|}{0.69} & 0.69 & \multicolumn{1}{c|}{11.91} & \multicolumn{1}{c|}{2.93} & 3.09 & \multicolumn{1}{c|}{15.46} & \multicolumn{1}{c|}{5.17} & 5.26 & \multicolumn{1}{c|}{17.24} & \multicolumn{1}{c|}{10.34} & 6.90 \\ \cline{4-16} 
 &  &  & VGen & \multicolumn{1}{c|}{14.71} & \multicolumn{1}{c|}{9.71} & 7.65 & \multicolumn{1}{c|}{\textbf{23.24}} & \multicolumn{1}{c|}{11.76} & 12.47 & \multicolumn{1}{c|}{\textbf{29.47}} & \multicolumn{1}{c|}{11.76} & 14.67 & \multicolumn{1}{c|}{\textbf{35.29}} & \multicolumn{1}{c|}{11.76} & 17.65 \\ \cline{3-16} 
 &  & \multirow{2}{*}{128K} & RTLLM & \multicolumn{1}{c|}{\textbf{5.52}} & \multicolumn{1}{c|}{0.52} & 0.34 & \multicolumn{1}{c|}{\textbf{14.58}} & \multicolumn{1}{c|}{2.40} & 1.54 & \multicolumn{1}{c|}{\textbf{19.82}} & \multicolumn{1}{c|}{4.36} & 2.63 & \multicolumn{1}{c|}{\textbf{27.59}} & \multicolumn{1}{c|}{6.90} & 3.45 \\ \cline{4-16} 
 &  &  & VGen & \multicolumn{1}{c|}{15.29} & \multicolumn{1}{c|}{9.71} & 9.41 & \multicolumn{1}{c|}{21.57} & \multicolumn{1}{c|}{11.76} & 11.74 & \multicolumn{1}{c|}{23.34} & \multicolumn{1}{c|}{11.76} & 11.76 & \multicolumn{1}{c|}{23.53} & \multicolumn{1}{c|}{11.76} & 11.76 \\ \hline
\multirow{16}{*}{Syntax} & \multirow{8}{*}{CodeLlama} & \multirow{2}{*}{34K} & RTLLM & \multicolumn{1}{c|}{60.52} & \multicolumn{1}{c|}{14.31} & 40.69 & \multicolumn{1}{c|}{80.61} & \multicolumn{1}{c|}{30.67} & 70.47 & \multicolumn{1}{c|}{84.52} & \multicolumn{1}{c|}{39.28} & 77.85 & \multicolumn{1}{c|}{86.21} & \multicolumn{1}{c|}{44.83} & 82.76 \\ \cline{4-16} 
 &  &  & VGen & \multicolumn{1}{c|}{86.76} & \multicolumn{1}{c|}{69.71} & 88.82 & \multicolumn{1}{c|}{99.14} & \multicolumn{1}{c|}{80.61} & 99.48 & \multicolumn{1}{c|}{99.97} & \multicolumn{1}{c|}{82.10} & 99.97 & \multicolumn{1}{c|}{\textbf{100.00}} & \multicolumn{1}{c|}{88.24} & \textbf{100.00} \\ \cline{3-16} 
 &  & \multirow{2}{*}{68K} & RTLLM & \multicolumn{1}{c|}{60.69} & \multicolumn{1}{c|}{26.90} & 53.45 & \multicolumn{1}{c|}{79.48} & \multicolumn{1}{c|}{51.12} & 75.87 & \multicolumn{1}{c|}{84.87} & \multicolumn{1}{c|}{62.85} & 81.10 & \multicolumn{1}{c|}{\textbf{89.66}} & \multicolumn{1}{c|}{68.97} & 82.76 \\ \cline{4-16} 
 &  &  & VGen & \multicolumn{1}{c|}{97.65} & \multicolumn{1}{c|}{71.76} & 60.88 & \multicolumn{1}{c|}{100.00} & \multicolumn{1}{c|}{82.40} & 89.90 & \multicolumn{1}{c|}{100.00} & \multicolumn{1}{c|}{86.48} & 93.73 & \multicolumn{1}{c|}{\textbf{100.00}} & \multicolumn{1}{c|}{88.24} & 94.12 \\ \cline{3-16} 
 &  & \multirow{2}{*}{102K} & RTLLM & \multicolumn{1}{c|}{\textbf{66.55}} & \multicolumn{1}{c|}{36.72} & 45.52 & \multicolumn{1}{c|}{\textbf{84.10}} & \multicolumn{1}{c|}{62.56} & 72.08 & \multicolumn{1}{c|}{\textbf{88.82}} & \multicolumn{1}{c|}{72.48} & 78.80 & \multicolumn{1}{c|}{\textbf{89.66}} & \multicolumn{1}{c|}{79.31} & 82.76 \\ \cline{4-16} 
 &  &  & VGen & \multicolumn{1}{c|}{96.47} & \multicolumn{1}{c|}{66.18} & 75.59 & \multicolumn{1}{c|}{\textbf{100.00}} & \multicolumn{1}{c|}{78.82} & 97.14 & \multicolumn{1}{c|}{\textbf{100.00}} & \multicolumn{1}{c|}{81.64} & 99.80 & \multicolumn{1}{c|}{\textbf{100.00}} & \multicolumn{1}{c|}{88.24} & \textbf{100.00} \\ \cline{3-16} 
 &  & \multirow{2}{*}{136K} & RTLLM & \multicolumn{1}{c|}{66.38} & \multicolumn{1}{c|}{39.48} & 33.28 & \multicolumn{1}{c|}{80.97} & \multicolumn{1}{c|}{61.11} & 64.90 & \multicolumn{1}{c|}{84.46} & \multicolumn{1}{c|}{67.88} & 74.84 & \multicolumn{1}{c|}{86.21} & \multicolumn{1}{c|}{72.41} & 79.31 \\ \cline{4-16} 
 &  &  & VGen & \multicolumn{1}{c|}{\textbf{99.12}} & \multicolumn{1}{c|}{67.65} & 73.53 & \multicolumn{1}{c|}{\textbf{100.00}} & \multicolumn{1}{c|}{79.89} & 96.94 & \multicolumn{1}{c|}{\textbf{100.00}} & \multicolumn{1}{c|}{82.65} & 99.78 & \multicolumn{1}{c|}{\textbf{100.00}} & \multicolumn{1}{c|}{88.24} & \textbf{100.00} \\ \cline{2-16} 
 & \multirow{8}{*}{CodeT5p} & \multirow{2}{*}{32K} & RTLLM & \multicolumn{1}{c|}{12.59} & \multicolumn{1}{c|}{9.66} & 10.86 & \multicolumn{1}{c|}{33.97} & \multicolumn{1}{c|}{22.50} & 32.60 & \multicolumn{1}{c|}{46.19} & \multicolumn{1}{c|}{28.52} & 45.12 & \multicolumn{1}{c|}{58.62} & \multicolumn{1}{c|}{34.48} & 51.72 \\ \cline{4-16} 
 &  &  & VGen & \multicolumn{1}{c|}{63.53} & \multicolumn{1}{c|}{17.06} & 34.12 & \multicolumn{1}{c|}{85.74} & \multicolumn{1}{c|}{47.09} & 71.55 & \multicolumn{1}{c|}{90.86} & \multicolumn{1}{c|}{59.36} & 83.75 & \multicolumn{1}{c|}{94.12} & \multicolumn{1}{c|}{70.59} & 94.12 \\ \cline{3-16} 
 &  & \multirow{2}{*}{64K} & RTLLM & \multicolumn{1}{c|}{28.28} & \multicolumn{1}{c|}{16.21} & 15.17 & \multicolumn{1}{c|}{52.27} & \multicolumn{1}{c|}{35.71} & 38.19 & \multicolumn{1}{c|}{61.96} & \multicolumn{1}{c|}{43.35} & 51.73 & \multicolumn{1}{c|}{68.97} & \multicolumn{1}{c|}{51.72} & 65.52 \\ \cline{4-16} 
 &  &  & VGen & \multicolumn{1}{c|}{74.12} & \multicolumn{1}{c|}{46.76} & 26.47 & \multicolumn{1}{c|}{86.85} & \multicolumn{1}{c|}{77.14} & 64.50 & \multicolumn{1}{c|}{88.14} & \multicolumn{1}{c|}{84.70} & 79.40 & \multicolumn{1}{c|}{88.24} & \multicolumn{1}{c|}{88.24} & 88.24 \\ \cline{3-16} 
 &  & \multirow{2}{*}{96K} & RTLLM & \multicolumn{1}{c|}{33.28} & \multicolumn{1}{c|}{12.76} & 24.83 & \multicolumn{1}{c|}{59.49} & \multicolumn{1}{c|}{29.05} & 46.58 & \multicolumn{1}{c|}{67.72} & \multicolumn{1}{c|}{38.94} & 59.44 & \multicolumn{1}{c|}{72.41} & \multicolumn{1}{c|}{48.28} & 68.97 \\ \cline{4-16} 
 &  &  & VGen & \multicolumn{1}{c|}{\textbf{82.35}} & \multicolumn{1}{c|}{42.94} & 52.65 & \multicolumn{1}{c|}{97.79} & \multicolumn{1}{c|}{72.81} & 83.97 & \multicolumn{1}{c|}{\textbf{99.93}} & \multicolumn{1}{c|}{81.68} & 92.84 & \multicolumn{1}{c|}{\textbf{100.00}} & \multicolumn{1}{c|}{88.24} & 94.12 \\ \cline{3-16} 
 &  & \multirow{2}{*}{128K} & RTLLM & \multicolumn{1}{c|}{\textbf{44.83}} & \multicolumn{1}{c|}{10.69} & 19.48 & \multicolumn{1}{c|}{\textbf{62.67}} & \multicolumn{1}{c|}{24.93} & 37.30 & \multicolumn{1}{c|}{\textbf{72.54}} & \multicolumn{1}{c|}{33.22} & 46.31 & \multicolumn{1}{c|}{\textbf{79.31}} & \multicolumn{1}{c|}{41.38} & 55.17 \\ \cline{4-16} 
 &  &  & VGen & \multicolumn{1}{c|}{77.35} & \multicolumn{1}{c|}{43.82} & 35.88 & \multicolumn{1}{c|}{\textbf{98.00}} & \multicolumn{1}{c|}{75.25} & 71.91 & \multicolumn{1}{c|}{99.89} & \multicolumn{1}{c|}{84.38} & 82.89 & \multicolumn{1}{c|}{\textbf{100.00}} & \multicolumn{1}{c|}{94.12} & 88.24 \\ \hline
\end{tabular}%
}
\label{tab:result}
\end{table*}

\subsection{Evaluation Benchmark and Metric}
We use RTLLM~\cite{lu2024rtllm} and VGen~\cite{thakur2024verigen} as evaluation benchmarks. Specifically, we employ low-level prompts from VGen that align with the format of our training data. These prompts describe the module's function along with its header, including the module name and the input and output types, which are the most challenging cases.

\subsubsection{Speed Evaluation}
In addition to prompts from RTLLM and VGen, we utilize GPT-4 to generate additional prompts for the Verilog code generation task based on the input prompt formats of RTLLM and VGen, aiming to enhance testing accuracy by increasing the diversity of prompts used during evaluation.
Ultimately, the generation speed of the fine-tuned models is assessed using a total of 575 input prompts.
For each prompt, the model generate outputs using both greedy decoding and sampling decoding methods, with the inference time recorded separately for each method. We then calculate the model’s generation speed using the following formula:
\begin{equation}
    \text{\textit{Speed}} = \frac{1}{n} \sum_{i=1}^{n} \frac{\text{Output Token Length}_i}{\text{Inference Time}_i}
    \label{eq:speed}
\end{equation}
where $n$ represents the total number of outputs generated (\textit{i.e.}, 575 outputs for each decoding method, resulting in $575\times2$ for the two decoding strategies).

The speedup of each fine-tuned model is calculated relative to its counterpart fine-tuned with NTP, which serves as the baseline, and is defined as follows:
\begin{equation}
    \text{\textit{Speedup}} =\frac{\text{Speed of Fine-tuned Model}}{\text{Speed of Fine-tuned Model with NTP}}
    \label{eq:speedup}
\end{equation}

\subsubsection{Quality Evaluation}



For syntax evaluation, a design is considered syntactically correct if both the design and its testbench successfully compile together using iverilog~\cite{iverilog}.
For functionality evaluation, a design is deemed functionally correct if its output matches the expected results for all testbench-provided stimuli.
We use the $pass$@$k$ metric, introduced in VerilogEval~\cite{liu2023verilogeval}, to evaluate both the functional and syntactic correctness of Verilog code generated by LLMs. For a specific prompt $i$, $pass$@$k$ reflects the likelihood of at least one correct solution among  $k$  randomly selected attempts: 
\begin{equation}
    pass\text{@}k=\underset{\text{\textit{prompt}}_i}{\mathbb{E}}\left[\frac{1-\binom{n-c}{k}}{\binom{n}{k}}\right]
    \label{eq:passk}
\end{equation}
Here, $n$ denotes the total number of samples generated by the model for each prompt, and $c$ represents the count of outputs that pass the functional check. To ensure a comprehensive assessment while maintaining experimental efficiency, we set $n = 20$ for all prompts and evaluate $k$ at values 1, 5, and 10.

\begin{figure}[ht]
    \centering
    \includegraphics[width=0.96\linewidth]{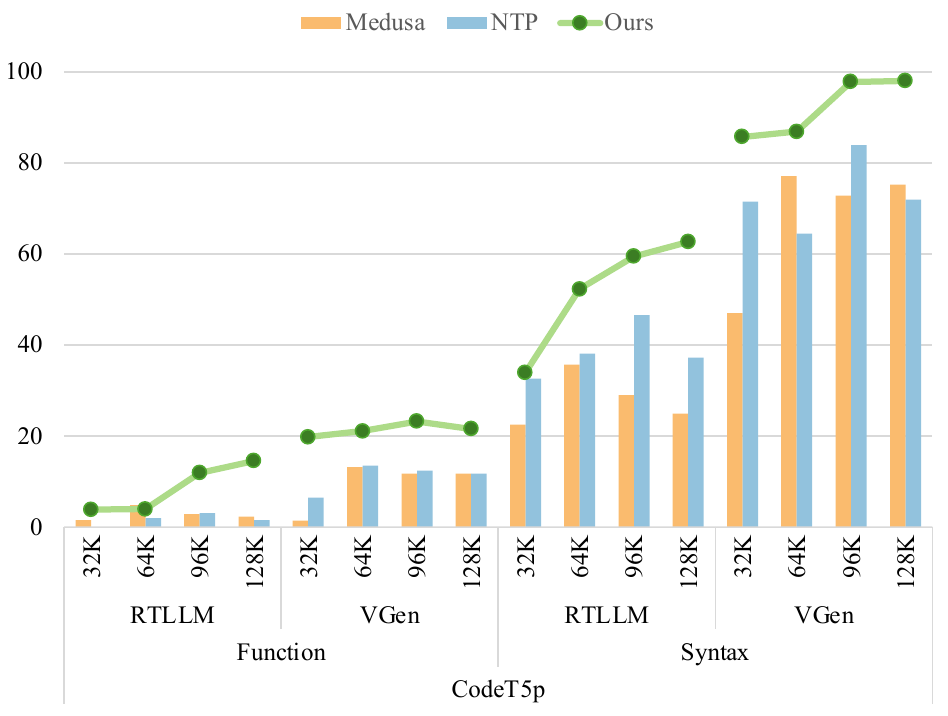}
    \caption{The comparison of pass@5 output quality between our method and baselines using the CodeT5p architecture.}
    \vspace{-6pt}
    \label{fig:compare_all}
\end{figure}

We also use an additional evaluation criterion, $\textit{Pass Rate}$, to further assess model performance. For each example in the benchmark, the model is considered successful if any of the 20 generated attempts passes validation. Given $m$ successful cases, the overall $\textit{Pass Rate}$ is calculated as:

\begin{equation}
\text{\textit{Pass Rate}} = \frac{m}{len(\text{Benchmark})}
\label{eq:passrate}
\end{equation}

\subsection{Experimental Result}

\begin{table}[ht]
\centering
\caption{Evaluation results for the speed of generating Verilog code}
\begin{tabular}{|c|cc|cc|}
\hline
\multirow{2}{*}{Method} & \multicolumn{2}{c|}{CodeLlama} & \multicolumn{2}{c|}{CodeT5p} \\ \cline{2-5} 
 & \multicolumn{1}{c|}{Speed (tokens/s)} & Speedup & \multicolumn{1}{c|}{Speed (tokens/s)} & Speedup \\ \hline
Ours & \multicolumn{1}{c|}{\textbf{420.13}} & \textbf{5.05} & \multicolumn{1}{c|}{\textbf{243.70}} & \textbf{2.66} \\ \hline
Medusa & \multicolumn{1}{c|}{294.99} & 3.55 & \multicolumn{1}{c|}{106.33} & 1.16 \\ \hline
NTP & \multicolumn{1}{c|}{83.13} & 1 & \multicolumn{1}{c|}{91.65} & 1 \\ \hline
\end{tabular}
\label{tab:speedup}
\end{table}

Table~\ref{tab:speedup} highlights the superior inference speed achieved by models fine-tuned with our method compared to those trained using Medusa and NTP. The most significant improvement is observed with the CodeLlama model, which achieves a speed of 420.13 tokens per second, corresponding to a 5.05× speedup over the NTP baseline. For CodeT5p, our method delivers a 2.66× speedup, outperforming the model trained with Medusa, which achieves a 1.16× speedup.
To further illustrate the effectiveness of our method, Fig.~\ref{fig:output_example} compares the decoding process of our approach to those of Medusa and NTP for a specific example.
Notably, our method generates the output in significantly fewer steps while maintaining the integrity of the syntactic structure at each decoding step.

The output quality of models fine-tuned with varying amounts of training data on the RTLLM~\cite{lu2024rtllm} and VGen~\cite{thakur2024verigen} benchmarks is summarized in Table~\ref{tab:result}. The highest values for each benchmark, within the same model architecture and across different training data sizes and methods, are highlighted in bold. 
For clearer visualization, Fig.~\ref{fig:compare_all} highlights the pass@5 results of our method compared to the baselines using the CodeT5p architecture.

The models trained with our method show significant improvements in both functional and syntactic accuracy compared to those trained with Medusa. On the VGen benchmark, our method achieves a maximum functional accuracy increase of 30.5\% using the pass@10 metric. For syntactic accuracy, it delivers a substantial improvement of 49.94\% in the pass@5 metric on the RTLLM benchmark.
When compared to the models fine-tuned with NTP, our approach achieves a functional accuracy improvement of up to 17.19\% using the pass@10 metric on the RTLLM benchmark and a syntactic accuracy gain of up to 47.65\% in the pass@1 metric on the VGen benchmark. Averaged across all benchmarks and evaluation metrics, models trained with our method show a 13.03\% improvement in functional accuracy over Medusa and a 5.91\% improvement over NTP. For syntactic accuracy, our method demonstrates an average enhancement of 22.9\% over Medusa and 11.8\% over NTP.
Additionally, our approach's strong performance on small datasets highlights its effectiveness, achieving competitive results without requiring extensive additional data.

\vspace{4pt}
\section{Conclusion}\label{Sec:Conclusion}
\vspace{4pt}
In this work, we introduce a novel application of speculative decoding for Verilog code generation, demonstrating its potential to enhance both inference speed and output quality. 
By aligning decoding stops with syntactically significant tokens extracted from ASTs, our method addresses the limitations of conventional tokenization and grammar-based approaches, enabling models to more effectively capture Verilog's structural and semantic intricacies. 
The proposed method achieves significant advancements in generating both syntactically and functionally correct code while substantially accelerating inference, delivering up to a 17.19\% improvement in pass@10 functional accuracy on the RTLLM benchmark and achieves up to a 5.05× speedup in Verilog code generation compared to the conventional NTP scheme. Furthermore, our approach enhances model inference speed by 1.42–2.29× over the original M\textsc{edusa} method.
These results underscore the efficacy of leveraging syntax-enriched speculative decoding to unlock new possibilities for applying LLMs to specialized programming languages, paving the way for more efficient and reliable Verilog code generation.


\section{Acknowledgments}

This work was supported in part by the General Research Fund of the Hong Kong Research Grants Council (RGC) under Grant No. 14212422 and 14202824, and in part by National Technology Innovation Center for EDA.

\balance
\newpage

\bibliographystyle{IEEEtran}
\bibliography{reference}

\begin{thebibliography}{10}
\providecommand{\url}[1]{#1}
\csname url@samestyle\endcsname
\providecommand{\newblock}{\relax}
\providecommand{\bibinfo}[2]{#2}
\providecommand{\BIBentrySTDinterwordspacing}{\spaceskip=0pt\relax}
\providecommand{\BIBentryALTinterwordstretchfactor}{4}
\providecommand{\BIBentryALTinterwordspacing}{\spaceskip=\fontdimen2\font plus
\BIBentryALTinterwordstretchfactor\fontdimen3\font minus \fontdimen4\font\relax}
\providecommand{\BIBforeignlanguage}[2]{{%
\expandafter\ifx\csname l@#1\endcsname\relax
\typeout{** WARNING: IEEEtran.bst: No hyphenation pattern has been}%
\typeout{** loaded for the language `#1'. Using the pattern for}%
\typeout{** the default language instead.}%
\else
\language=\csname l@#1\endcsname
\fi
#2}}
\providecommand{\BIBdecl}{\relax}
\BIBdecl

\bibitem{wang2021codet5}
Y.~Wang, W.~Wang, S.~Joty, and S.~C. Hoi, ``Codet5: Identifier-aware unified pre-trained encoder-decoder models for code understanding and generation,'' in \emph{Proceedings of the 2021 Conference on Empirical Methods in Natural Language Processing}, 2021, pp. 8696--8708.

\bibitem{wang2023codet5+}
Y.~Wang, H.~Le, A.~Gotmare, N.~Bui, J.~Li, and S.~Hoi, ``Codet5+: Open code large language models for code understanding and generation,'' in \emph{Proceedings of the 2023 Conference on Empirical Methods in Natural Language Processing}, 2023, pp. 1069--1088.

\bibitem{roziere2023code}
B.~Roziere, J.~Gehring, F.~Gloeckle, S.~Sootla, I.~Gat, X.~E. Tan, Y.~Adi, J.~Liu, R.~Sauvestre, T.~Remez \emph{et~al.}, ``Code llama: Open foundation models for code,'' \emph{arXiv preprint arXiv:2308.12950}, 2023.

\bibitem{zhu2024deepseek}
Q.~Zhu, D.~Guo, Z.~Shao, D.~Yang, P.~Wang, R.~Xu, Y.~Wu, Y.~Li, H.~Gao, S.~Ma \emph{et~al.}, ``Deepseek-coder-v2: Breaking the barrier of closed-source models in code intelligence,'' \emph{arXiv preprint arXiv:2406.11931}, 2024.

\bibitem{sennrich2015neural}
R.~Sennrich, ``Neural machine translation of rare words with subword units,'' \emph{arXiv preprint arXiv:1508.07909}, 2015.

\bibitem{rabinovich2017abstract}
M.~Rabinovich, M.~Stern, and D.~Klein, ``Abstract syntax networks for code generation and semantic parsing,'' in \emph{Proceedings of the 55th Annual Meeting of the Association for Computational Linguistics (Volume 1: Long Papers)}, 2017, pp. 1139--1149.

\bibitem{sun2019grammar}
Z.~Sun, Q.~Zhu, L.~Mou, Y.~Xiong, G.~Li, and L.~Zhang, ``A grammar-based structural cnn decoder for code generation,'' in \emph{Proceedings of the AAAI conference on artificial intelligence}, vol.~33, no.~01, 2019, pp. 7055--7062.

\bibitem{sun2020treegen}
Z.~Sun, Q.~Zhu, Y.~Xiong, Y.~Sun, L.~Mou, and L.~Zhang, ``Treegen: A tree-based transformer architecture for code generation,'' in \emph{Proceedings of the AAAI conference on artificial intelligence}, vol.~34, no.~05, 2020, pp. 8984--8991.

\bibitem{xiong2022l2s}
Y.~Xiong and B.~Wang, ``L2s: A framework for synthesizing the most probable program under a specification,'' \emph{ACM Transactions on Software Engineering and Methodology (TOSEM)}, vol.~31, no.~3, pp. 1--45, 2022.

\bibitem{zhu2022grape}
Q.~Zhu, Z.~Sun, W.~Zhang, Y.~Xiong, and L.~Zhang, ``Grape: Grammar-preserving rule embedding.'' in \emph{IJCAI}, 2022, pp. 4545--4551.

\bibitem{zhu2024grammart5}
Q.~Zhu, Q.~Liang, Z.~Sun, Y.~Xiong, L.~Zhang, and S.~Cheng, ``Grammart5: Grammar-integrated pretrained encoder-decoder neural model for code,'' in \emph{Proceedings of the IEEE/ACM 46th International Conference on Software Engineering}, 2024, pp. 1--13.

\bibitem{leviathan2023fast}
Y.~Leviathan, M.~Kalman, and Y.~Matias, ``Fast inference from transformers via speculative decoding,'' in \emph{International Conference on Machine Learning}.\hskip 1em plus 0.5em minus 0.4em\relax PMLR, 2023, pp. 19\,274--19\,286.

\bibitem{chen2023accelerating}
C.~Chen, S.~Borgeaud, G.~Irving, J.-B. Lespiau, L.~Sifre, and J.~Jumper, ``Accelerating large language model decoding with speculative sampling,'' \emph{arXiv preprint arXiv:2302.01318}, 2023.

\bibitem{miao2024specinfer}
X.~Miao, G.~Oliaro, Z.~Zhang, X.~Cheng, Z.~Wang, Z.~Zhang, R.~Y.~Y. Wong, A.~Zhu, L.~Yang, X.~Shi \emph{et~al.}, ``Specinfer: Accelerating large language model serving with tree-based speculative inference and verification,'' in \emph{Proceedings of the 29th ACM International Conference on Architectural Support for Programming Languages and Operating Systems, Volume 3}, 2024, pp. 932--949.

\bibitem{cai2024medusa}
T.~Cai, Y.~Li, Z.~Geng, H.~Peng, J.~D. Lee, D.~Chen, and T.~Dao, ``Medusa: Simple llm inference acceleration framework with multiple decoding heads,'' \emph{arXiv preprint arXiv:2401.10774}, 2024.

\bibitem{lu2024rtllm}
Y.~Lu, S.~Liu, Q.~Zhang, and Z.~Xie, ``Rtllm: An open-source benchmark for design rtl generation with large language model,'' in \emph{2024 29th Asia and South Pacific Design Automation Conference (ASP-DAC)}.\hskip 1em plus 0.5em minus 0.4em\relax IEEE, 2024, pp. 722--727.

\bibitem{pearce2020dave}
H.~Pearce, B.~Tan, and R.~Karri, ``Dave: Deriving automatically verilog from english,'' in \emph{Proceedings of the 2020 ACM/IEEE Workshop on Machine Learning for CAD}, 2020, pp. 27--32.

\bibitem{blocklove2023chip}
J.~Blocklove, S.~Garg, R.~Karri, and H.~Pearce, ``Chip-chat: Challenges and opportunities in conversational hardware design,'' in \emph{2023 ACM/IEEE 5th Workshop on Machine Learning for CAD (MLCAD)}.\hskip 1em plus 0.5em minus 0.4em\relax IEEE, 2023, pp. 1--6.

\bibitem{thakur2023benchmarking}
S.~Thakur, B.~Ahmad, Z.~Fan, H.~Pearce, B.~Tan, R.~Karri, B.~Dolan-Gavitt, and S.~Garg, ``Benchmarking large language models for automated verilog rtl code generation,'' in \emph{2023 Design, Automation \& Test in Europe Conference \& Exhibition (DATE)}.\hskip 1em plus 0.5em minus 0.4em\relax IEEE, 2023, pp. 1--6.

\bibitem{liu2023verilogeval}
M.~Liu, N.~Pinckney, B.~Khailany, and H.~Ren, ``Verilogeval: Evaluating large language models for verilog code generation,'' in \emph{2023 IEEE/ACM International Conference on Computer Aided Design (ICCAD)}.\hskip 1em plus 0.5em minus 0.4em\relax IEEE, 2023, pp. 1--8.

\bibitem{chang2023chipgpt}
K.~Chang, Y.~Wang, H.~Ren, M.~Wang, S.~Liang, Y.~Han, H.~Li, and X.~Li, ``Chipgpt: How far are we from natural language hardware design,'' \emph{arXiv preprint arXiv:2305.14019}, 2023.

\bibitem{thakur2024verigen}
S.~Thakur, B.~Ahmad, H.~Pearce, B.~Tan, B.~Dolan-Gavitt, R.~Karri, and S.~Garg, ``Verigen: A large language model for verilog code generation,'' \emph{ACM Transactions on Design Automation of Electronic Systems}, vol.~29, no.~3, pp. 1--31, 2024.

\bibitem{liu2024rtlcoder}
S.~Liu, W.~Fang, Y.~Lu, Q.~Zhang, H.~Zhang, and Z.~Xie, ``Rtlcoder: Outperforming gpt-3.5 in design rtl generation with our open-source dataset and lightweight solution,'' in \emph{2024 IEEE LLM Aided Design Workshop (LAD)}.\hskip 1em plus 0.5em minus 0.4em\relax IEEE, 2024, pp. 1--5.

\bibitem{chang2024data}
K.~Chang, K.~Wang, N.~Yang, Y.~Wang, D.~Jin, W.~Zhu, Z.~Chen, C.~Li, H.~Yan, Y.~Zhou \emph{et~al.}, ``Data is all you need: Finetuning llms for chip design via an automated design-data augmentation framework,'' in \emph{Proceedings of the 61st ACM/IEEE Design Automation Conference}, 2024, pp. 1--6.

\bibitem{zhang2024mg}
Y.~Zhang, Z.~Yu, Y.~Fu, C.~Wan, and Y.~C. Lin, ``Mg-verilog: Multi-grained dataset towards enhanced llm-assisted verilog generation,'' in \emph{2024 IEEE LLM Aided Design Workshop (LAD)}.\hskip 1em plus 0.5em minus 0.4em\relax IEEE, 2024, pp. 1--5.

\bibitem{cui2024origen}
F.~Cui, C.~Yin, K.~Zhou, Y.~Xiao, G.~Sun, Q.~Xu, Q.~Guo, D.~Song, D.~Lin, X.~Zhang \emph{et~al.}, ``Origen: Enhancing rtl code generation with code-to-code augmentation and self-reflection,'' \emph{arXiv preprint arXiv:2407.16237}, 2024.

\bibitem{zhao2024codev}
Y.~Zhao, D.~Huang, C.~Li, P.~Jin, Z.~Nan, T.~Ma, L.~Qi, Y.~Pan, Z.~Zhang, R.~Zhang \emph{et~al.}, ``Codev: Empowering llms for verilog generation through multi-level summarization,'' \emph{arXiv preprint arXiv:2407.10424}, 2024.

\bibitem{wang2021syncobert}
X.~Wang, Y.~Wang, F.~Mi, P.~Zhou, Y.~Wan, X.~Liu, L.~Li, H.~Wu, J.~Liu, and X.~Jiang, ``Syncobert: Syntax-guided multi-modal contrastive pre-training for code representation,'' \emph{arXiv preprint arXiv:2108.04556}, 2021.

\bibitem{jiang2021treebert}
X.~Jiang, Z.~Zheng, C.~Lyu, L.~Li, and L.~Lyu, ``Treebert: A tree-based pre-trained model for programming language,'' in \emph{Uncertainty in Artificial Intelligence}.\hskip 1em plus 0.5em minus 0.4em\relax PMLR, 2021, pp. 54--63.

\bibitem{guo2022unixcoder}
D.~Guo, S.~Lu, N.~Duan, Y.~Wang, M.~Zhou, and J.~Yin, ``Unixcoder: Unified cross-modal pre-training for code representation,'' in \emph{Proceedings of the 60th Annual Meeting of the Association for Computational Linguistics (Volume 1: Long Papers)}, 2022, pp. 7212--7225.

\bibitem{yan2017privmin}
Z.~Yan, J.~Liu, G.~Li, Z.~Han, and S.~Qiu, ``Privmin: Differentially private minhash for jaccard similarity computation,'' \emph{arXiv preprint arXiv:1705.07258}, 2017.

\bibitem{chen2023incremental}
X.~Chen, Y.~Meng, and G.~Chen, ``Incremental verilog parser,'' in \emph{2023 International Symposium of Electronics Design Automation (ISEDA)}.\hskip 1em plus 0.5em minus 0.4em\relax IEEE, 2023, pp. 236--240.

\bibitem{achiam2023gpt}
J.~Achiam, S.~Adler, S.~Agarwal, L.~Ahmad, I.~Akkaya, F.~L. Aleman, D.~Almeida, J.~Altenschmidt, S.~Altman, S.~Anadkat \emph{et~al.}, ``Gpt-4 technical report,'' \emph{arXiv preprint arXiv:2303.08774}, 2023.

\bibitem{alpaca}
R.~Taori, I.~Gulrajani, T.~Zhang, Y.~Dubois, X.~Li, C.~Guestrin, P.~Liang, and T.~B. Hashimoto, ``Stanford alpaca: An instruction-following llama model,'' \url{https://github.com/tatsu-lab/stanford_alpaca}, 2023.

\bibitem{axolotl2023}
Axolotl, ``{Axolotl},'' \url{https://github.com/OpenAccess-AI-Collective/axolotl}, 2023.

\bibitem{dettmers2023qlora}
T.~Dettmers, A.~Pagnoni, A.~Holtzman, and L.~Zettlemoyer, ``Qlora: Efficient finetuning of quantized llms,'' \emph{arXiv preprint arXiv:2305.14314}, 2023.

\bibitem{iverilog}
S.~Williams, ``Icarus verilog,'' \url{http://iverilog.icarus.com}, 2024, accessed: 2024-11-19.

\end{thebibliography}

\end{document}